\documentclass[10pt,twocolumn,letterpaper]{article}

\usepackage{cvpr}              % To produce the CAMERA-READY version
\usepackage{adjustbox}
\usepackage{array}
\newcolumntype{R}[2]{%
    >{\adjustbox{angle=#1,lap=1.3\width-(#2)}\bgroup}%
    l%
    <{\egroup}%
}

\usepackage{booktabs}
\usepackage{multirow}

\usepackage{soul}

\usepackage[dvipsnames]{xcolor}

\definecolor{cvprblue}{rgb}{0.21,0.49,0.74}
\usepackage[pagebackref,breaklinks,colorlinks,citecolor=cvprblue]{hyperref}
\usepackage{graphicx}
\usepackage{amsmath}
\usepackage{amssymb}
\usepackage{amsfonts}
\usepackage{booktabs}
\usepackage{color, colortbl}
\usepackage{pifont}
\usepackage{mathrsfs}
\usepackage{makecell}
\usepackage{xcolor} 
\definecolor{LightCyan}{rgb}{0.88,1,1}
\definecolor{better}{rgb}{0.19, 0.55, 0.91}
\definecolor{worse}{rgb}{0.82, 0.1, 0.26}

\def\paperID{1766} % *** Enter the Paper ID here
\def\confName{CVPR}
\def\confYear{2024}

\title{Building a Strong Pre-Training Baseline for Universal 3D Large-Scale Perception}

\author{
Haoming Chen$\textsuperscript{1}$, 
Zhizhong Zhang{$\textsuperscript{1}$}{$\textsuperscript{\dag}$}, 
Yanyun Qu$\textsuperscript{3}$, 
Ruixin Zhang$\textsuperscript{4}$, 
Xin Tan$\textsuperscript{1,2}$, 
Yuan Xie$\textsuperscript{1,2}$\\
$\textsuperscript{1}$East China Normal University, Shanghai, China\\
$\textsuperscript{2}$Chongqing Institute of East China Normal University, Chongqing, China \\
$\textsuperscript{3}$Xiamen University, Fujian, China\\
$\textsuperscript{4}$Tencent Youtu Lab, Shanghai, China \\
\tt\small
chenhaomingbob@gmail.com, \{zzzhang, xtan, yxie\}@cs.ecnu.edu.cn \\
\tt\small
yyqu@xmu.edu.cn, ruixinzhang@tencent.com \\
$\textsuperscript{\dag}$
\tt\small
Corresponding Author
}

\begin{document}
% \maketitle

\twocolumn[
{
\renewcommand\twocolumn[1][]{#1}%
\maketitle
\vspace{-4em}
\begin{center}
    
    \includegraphics[width=\textwidth]{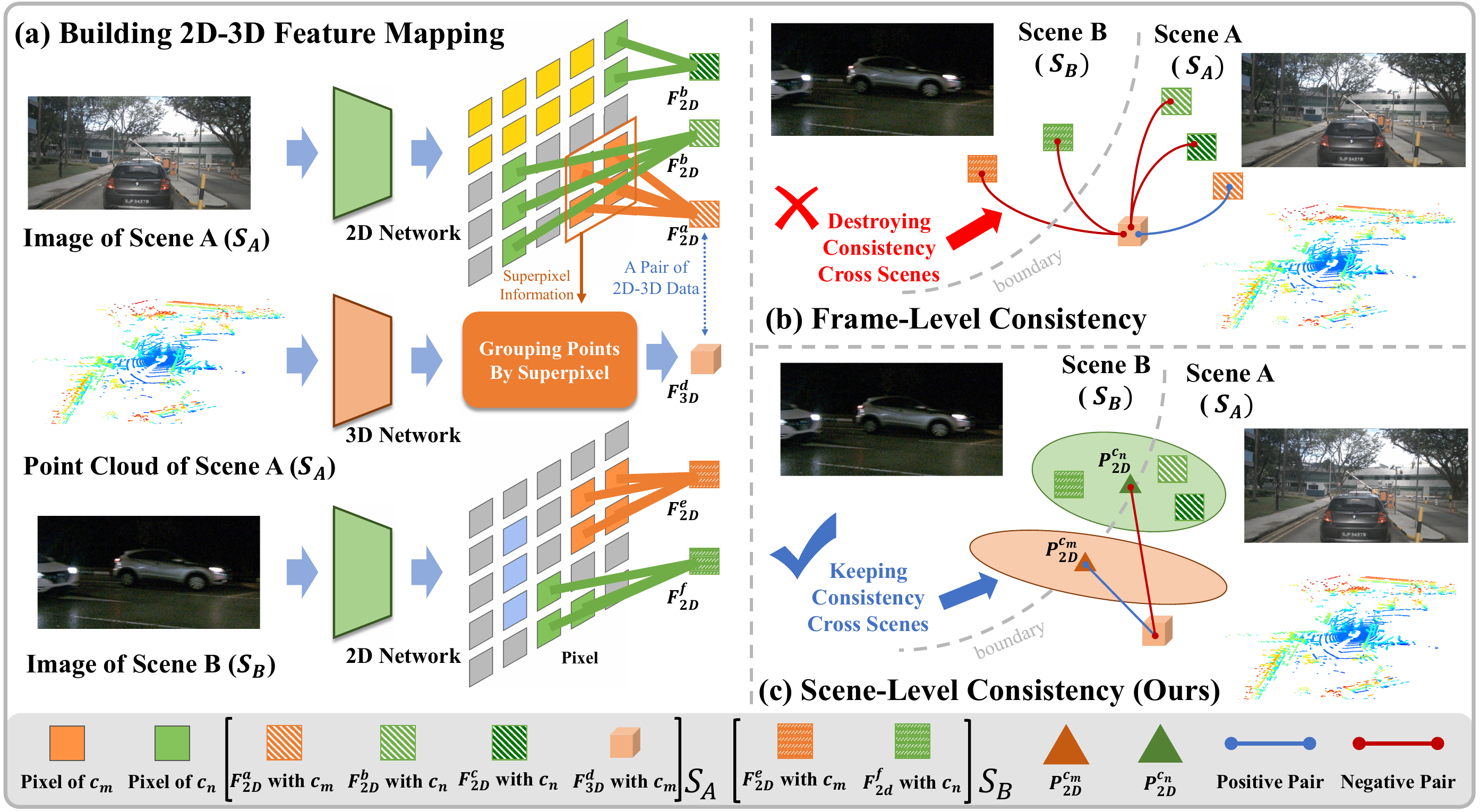}    
    \vspace{-2.0em}
    \captionof{figure}{ Brief illustration of the current multi-modality 3D pre-training paradigm and our proposed scene-level consistency. (a) we show the process of superpixel-superpoint association and clearly observe that superpixels with the same semantics can exist in the same scene or in different scenes, \emph{e.g.}, the green squares. (b) We summarize existing pre-training methods and find that they all use frame-level consistency to learn 3D representations. Moreover, we believe that this constraint breaks the semantic consistency across views/frames and visualize the drawback. (c) We show our proposed scene-level consistency, which keeping the semantic consistency across various scenes. As a result, we achieve SOTA on three perceptual tasks with limited 3D annotation (Tab. \ref{tab:three_ann_eff_tasks}). Our CSC framework builds a strong pre-training baseline for universal 3D Large-scale perception.
    }   
    \label{fig:motivation}
    % \vspace{-1.0em}
\end{center}
}
]
% \footnote{This is the footnote text.}

% \footnote{\thanks{Corresponding Authors}}

\begin{abstract}
\vspace{-1.0em}
An effective pre-training framework with universal 3D representations is extremely desired in perceiving large-scale dynamic scenes. However, establishing such an ideal framework that is both task-generic and label-efficient poses a challenge in unifying the representation of the same primitive across diverse scenes. The current contrastive 3D pre-training methods typically follow a frame-level consistency, which focuses on the 2D-3D relationships in each detached image. Such inconsiderate consistency greatly hampers the promising path of reaching an universal pre-training framework: (1) The cross-scene semantic self-conflict, {\textit i.e.}, the intense collision between primitive segments of the same semantics from different scenes; (2) Lacking a globally unified bond that pushes the cross-scene semantic consistency into 3D representation learning. To address above challenges, we propose a CSC framework that puts a scene-level semantic consistency in the heart, bridging the connection of the similar semantic segments across various scenes. To achieve this goal, we combine the coherent semantic cues provided by the vision foundation model and the knowledge-rich cross-scene prototypes derived from the complementary multi-modality information. These allow us to train a universal 3D pre-training model that facilitates various downstream tasks with less fine-tuning efforts. Empirically, we achieve consistent improvements over SOTA pre-training approaches in semantic segmentation (+1.4\% mIoU), object detection (+1.0\% mAP), and panoptic segmentation (+3.0\% PQ) using their task-specific 3D network on nuScenes. Code is released at \href{https://github.com/chenhaomingbob/CSC}{https://github.com/chenhaomingbob/CSC}, hoping to inspire future research.

% large-scale dynamic scenes to allow each agent could running safely inside the 3D environment.
% 3D representation learning. Through tackle these problems, we build a strong pre-training baseline for universal 3D large-scale perception. Our framework put the cross-scenes semantic consistency at the heart. We achieve this idea by coupling the coherent 2D semantic cues and knowledge-rich unified prototypes shared in various scenes. These allow our framework is better than the current state-of-the-art  pre-training methods in all three important perception tasks. On benchmark dataset nuScenes, we achieve consistent boosts in semantic segmentation, object detection, and panoptic segmentation with task-specific 3D network.
 % Due to the nature of data hungry, existing supervised methods require a massive of high-quality 3D annotation via expensive cost

\end{abstract}    

% \vspace{-1.0em}
\section{Introduction}
\label{sec:introduction}
As one of the most promising applications of artificial intelligence, autonomous driving has undergone rapid development in recent years \cite{hu2023planning,li2022bevformer,yang2023bevformer,kong2023lasermix,kong2023robo3d}. In it, 3D scene perception plays a fundamental role to perceive and understand surroundings, and thus has become increasingly attended in recent researches \cite{kong2023lasermix,kong2023robo3d,liu2023uniseg,gong2022optimization}. Within the 3D perception, there are three essential tasks across different granularities, \textit{i.e.,} semantic segmentation \cite{riz2023novel,li2023less,li2023mseg3d,lai2023spherical,TPAMI_mqh,gong2021boundary,gong2021omni}, object detection \cite{wang2023frustumformer,feng2023aedet,klingner2023x3kd,jiao2023msmdfusion}, and panoptic segmentation \cite{Zhang_2023_ICCV,zhou2021panoptic,li2022panoptic}. All of them build upon powerful 3D representations and recent success  of vision foundation model shows great potential for such fabulous goal \cite{pang2023unsupervised,sautier2022image,liu2023segment,clip2scene,openscene}. 

% Even though supervised algorithms have achieved inspirational results, they still require large amounts of high-quality 3D annotation via expensive cost. Therefore, studies on 3D representation learning from the accessible and mature 2D vision attract lots of attention .

% 介绍2D图像对3D点云表征学习的帮助, 同时引入2D领域现在如火如荼的视觉大模型。
Thanks to the paired image-lidar data captured by the multi-sensors \cite{caesar2020nuscenes,behley2019semantickitti,sun2020scalability,geiger2012we}, it lays the foundation for improving 3D representation through cross-modality learning from 2D image priors. However, how VFM knowledge could benefit 3D scene perception without using any or limited point cloud annotations remains under-explored.

% Coincidently, the emerging paradigm of vision foundation models ({\it i.e.}, VFMs), where models are pre-trained on broad image data, has great potential in playing this role.

% 开始介绍现有工作和存在的缺陷
Inspired by the groundbreaking work SLidR \cite{sautier2022image}, a succession of multi-modality 3D self-supervised approaches \cite{mahmoud2023self,pang2023unsupervised,liu2023segment} has been proposed subsequently. To obtain desired 3D representation, all of them adopt a \textbf{frame-level consistency}, which conduct superpixel-superpoint contrastive distillation from the association between a point cloud frame and a image. We show the  2D-3D association in Fig. \ref{fig:motivation} (a).  We observe that achieving the pre-trained 3D backbone with strong generalization ability still faces the following \textbf{two challenges}: \textbf{(1)} the superpixels sharing the identical semantic from the same or different images are erroneously treated as the negative pair. Despite the introduction of VFMs \cite{liu2023segment} or the adoption of semantically tolerant loss \cite{mahmoud2023self} as coping strategies, the challenge remains unresolved, especially in the presence of numerous identical semantic superpixels observed across various views or scenes. As presented in Fig. \ref{fig:motivation} (b), the cross-scene superpixels ($\boldsymbol{F_{2D}^{a}}$ \& $\boldsymbol{F_{2D}^{e}}$) with the same semantic $\boldsymbol{\mathcal{C}}_{sem}^{m}$ are not consistent in these methods. \textbf{(2)} The challenge of keeping global semantic consistency in large-scale scenarios. As an example, for objects with identical semantic labels but from different frames, their features should be as close as possible, whether the frames are from the same or disparate scenes. 

In this paper, we introduce a strong pre-trained baseline, termed CSC (\textbf{\underline{C}}oherent \textbf{\underline{S}}emantic \textbf{\underline{C}}ues Framework), for universal 3D large-scale perception by learning the \textbf{scene-level consistency} (as shown in Fig. \ref{fig:motivation} (c)). The core idea is that we push the cross-scenes semantic consistency into the pristine 3D backbone, leveraging the coherent semantic cues provided by powerful VFMs and information-rich semantic prototypes from multi-modality. Specifically, CSC consists of two key components: \textbf{(i)} A VFM-Assisted Semantic Prototype Generation, where we first utilize the VFM to provide reliable and coherent semantic cues for all superpixels across diverse scenes and then produce the multi-modality semantic prototypes to cover the representative features of involved semantics. \textbf{(ii)} A Coherent Semantic Consistency for 3D universal representation learning. In this component, we propose a multi-modality prototype blending module designed to fuse these unaligned prototypes that, while semantically aligned, reside in distinct feature spaces. We individually process each prototypes through a modality-specific prototype projection module, yielding implicitly aligned prototypes. Subsequently, these updated prototypes are combined and fed into a multi-modality prototype fusion module, resulting in the mixed prototypes that incorporate information from both image and lidar modalities. Ultimately, we perform a cross-scene semantic contrastive loss between superpoints and  mixed prototypes, thereby obtaining the universal 3D representation with scene-level semantic consistency.

Compared to the current methods, the CSC brings a new SOTA performance to all three mainstream 3D perception tasks, semantic segmentation, object detection, and  panoptic segmentation. This is particularly notable when only limited 3D annotations are available for a specific task. The main contributions of this work are summarized as follows:
\begin{itemize}
    \item To the best of our knowledge, CSC is the first work to explore cross-scene semantic consistency in multi-modality 3D pre-training, which achieves the semantic consistency of all frames from all scenes.
    % and is beneficial to multiple perception tasks on large-scale 3D point clouds.
    % the proposed multi-modality 3D pre-training framework CSC is the first work to explore cross-scene semantic consistency assisted by the VFM, and is beneficial to multiple perception tasks on large-scale 3D point clouds.
    \item We utilize unexplored semantic cues from the VFM to maintain the coherent semantic prototypes including multi-modality information, resulting in the general pre-trained 3D backbone.
    
    % In order to learn unified 3D representations throughout all scenes, we utilize unexplored semantic cues from the VFM to maintain the coherent semantic prototypes including multi-modality information, resulting in the general pre-trained 3D backbone.
    \item Our CSC establishes a strong pre-training baseline for universal 3D large-scale perception, surpassing prior self-supervised approaches, remarkably evidenced by three annotation-efficient downstream tasks, semantic segmentation improved by 1.4\% mIoU, object detection by 1.0\% mAP, and panoptic segmentation by 3\% PQ. 
\end{itemize}

% \vspace{-1.0em}
\section{Related Works}
\label{sec:related_works}

% \vspace{-0.5em}
\subsection{Vision Foundation Models}
% \vspace{-0.5em}
In light of the advancement of massive and diverse image sources and inspired by large-scale vision-language pre-training techniques, the computer vision community \cite{OneFormer,SAM,DINOv2,liu2021deep,Feng_2023_CVPR, li2023vs, kong2022compactness} is now witnessing hot attention  in building powerful vision systems. Among these vision foundation models, DINOV2 and SAM have gathered the most widespread attention. The DINOv2 \cite{DINOv2} is an advanced self-supervised learning framework that leverages vision transformers (with 1B parameters) and enough curated data sources (about 142M images) to producing high-performance visual features. The segment anything model (SAM) \cite{SAM} demonstrates a new paradigm with robust zero-shot transferability, excelling in new image distributions and tasks. Apart from the mentioned works, approaches including SEEM \cite{SEEM} and OneFormer \cite{OneFormer} also expand the visual model landscape, offering alternatives for the vision community. In this study, we explore the potential of VFMs for universal 3D scene perception. We leverage the VFM to acquire reliable and stable semantic cues across images from diverse scenes, harnessing these cues to promote global semantic consistency learning of 3D representations.

\vspace{-1em}
\subsection{Self-Supervised 3D Representation Learning}
% \vspace{-0.5em}
Here, we focus on the line of contrastive-based self-supervised methods \cite{boulch2023also,wu2023spatiotemporal,sautier2023bevcontrast,sautier2022image,mahmoud2023self,pang2023unsupervised,liu2023segment,clip2scene,openscene,TARL}. According to the types of input modalities, these methods can be further categorized into uni-modality and multi-modality self-supervised framework. For uni-modality, they commonly perform the multi-view consistency constraint, which seek to the consistency of points/regions from various view transformations. TARL \cite{TARL} exploits vehicle motion to extract different views of the same object in consecutive point cloud frames to learn spatio-temporal view-consistent. When we pay attention to the point cloud acquisition, we will find that almost most of the point cloud data will have its corresponding image data. For multi-modality, current studies not only consider 2D images but also involve text. In this paper, we primarily discuss the assistance of 2D images to advance 3D perception. SLidR \cite{sautier2022image} is the pioneering work that take superpixels obtained from SLIC \cite{SLIC} as units, and achieves superpixel-driven contrastive distillation to initialize the 3D network. Subsequently, Seal \cite{liu2023segment}, the current SOTA method, introduces the popular VFM into this field and gains breakthrough performance improvements. Compared to these methods, we propose semantic prototype to manage the coherent semantic cues of all visual inputs, and enable scene-level semantic consistency to promote 3D representations.

% Similar to Seal, our work also utilize the VFM to produce superpixels, but differs from Seal in that we manage the coherent semantic cues of all visual inputs and further perform scene-level semantic consistency to promote 3D representations.

\subsection{Prototype-based Self-Supervised Learning}
In 2D self-supervised realm, a wide range of applications adopts the idea of clustering or prototype. In the USL-VI-ReID \cite{wang2022optimal}, the existing SOTA methods \cite{cclnet2023} are developed on the ClusterContrast \cite{clustercontrast}, which first generates prototypes using a clustering algorithm and then optimizes their networks via a cluster comparison mechanism. Meanwhile, in the field of unsupervised semantic segmentation, many excellent works \cite{wen2022self,li2023acseg,Luo2023SegCLIP} use the concept of prototypes and show inspiring performance. Inspired by these promising study, our CSC use prototypes for 3D pre-training. Through the prototypes, we could bridge the connection of various superpixels from different views or scenes.

% \vspace{-1.0em}
\section{Methodology}
\label{sec:methodology}

\begin{figure*}[t]
\centering
\includegraphics[width=1.0\linewidth]{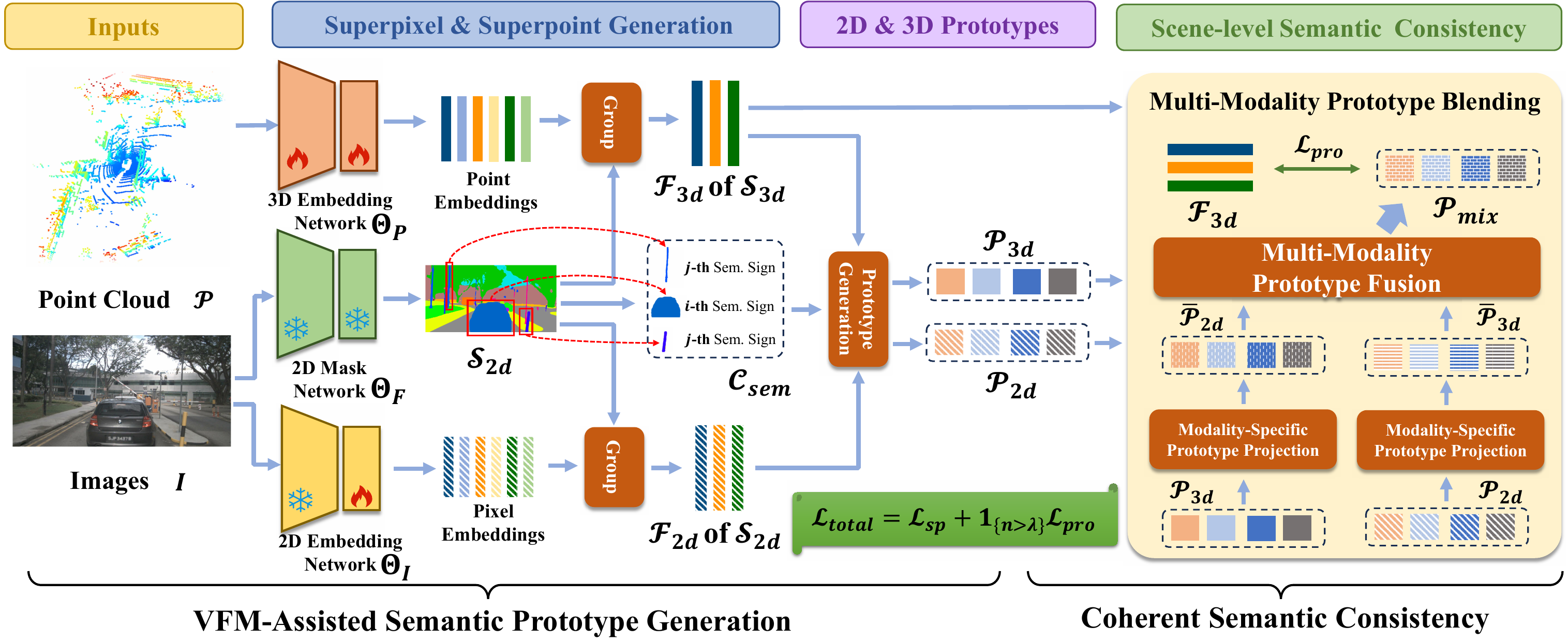}
\vspace{-1.5em}
\caption{ Overview of the CSC framework. CSC leverages the scene-level semantic consistency to obtain the universal 3D representations (Sec. \ref{sec:methodology}), and then fine-tunes the pre-trained 3D backbone for three downstream perception tasks (Sec. \ref{sec:experiments}). To achieve the scene-level semantic consistency, CSC consists of the VFM-assisted semantic prototype generation module (Sec. \ref{sec:2d_3d_prototypes}) and the coherent semantic consistency module (Sec. \ref{sec:prototype-based-Constraint}).}
\label{fig:framework}
\vspace{-0.5em}
\end{figure*}

\subsection{Preliminaries}
To learn powerful 3D representations, the current 3D pre-training paradigm (stemming from SLidR \cite{sautier2022image}) utilizes the calibrated relationships between 2D images and 3D point clouds with the assistance of powerful 2D models. Technically, for a point cloud frame $\boldsymbol{\mathcal{P}}=\left\{\mathbf{P}_k \mid k=1,\ldots, K  \right\}$ comprising $K$ points, each point $ \mathbf{P}_k \in \mathbb{R}^{4}$ represents the $k$-th point's three-dimensional location $(x,y,z)$ coupled with its intensity feature. Meanwhile, the point cloud frame is equipped with $L$ surrounding images $\boldsymbol{\mathcal{I}}=\left\{\mathbf{I}_l \mid l=1,\dots, L\right\}$, where $\mathbf{I}_l \in \mathbb{R}^{H \times W \times 3}$ denotes $l$-th RGB image with the shape of $H$ $\times$ $W$.

In the pre-training phase, the images $\boldsymbol{\mathcal{I}}$ and points $\boldsymbol{\mathcal{P}}$ are respectively fed into a 2D embedding network, $\boldsymbol{\Theta_I}: \mathbb{R}^{H \times W \times 3} \rightarrow \mathbb{R}^{H \times W \times D}$, and a 3D embedding network, $\boldsymbol{\Theta_P}: \mathbb{R}^{K \times 4} \rightarrow \mathbb{R}^{K \times D}$, to produce pixel-wise features and point-wise features. Next, grouping these features via 2D masks $\boldsymbol{\mathcal{S_{\text{2D}}}}$ (\emph{a.k.a}, superpixels introduced in \cite{sautier2022image}) obtained from a 2D segmentation algorithm, $\boldsymbol{\Theta_F}: \mathbb{R}^{H \times W \times 3} \rightarrow \mathbb{R}^{H \times W \times 1} $, we can harvest superpixel embeddings $\boldsymbol{\mathcal{F_{\text{2D}}}}$ and superpoint embeddings $\boldsymbol{\mathcal{F_{\text{3D}}}}$. Finally, the existing state-of-the-art methods \cite{liu2023segment,mahmoud2023self} will conduct frame-level superpixel-superpoint contrastive loss by utilizing the 2D-3D mapping, which pulls the matched superpixel-superpoint features while pushing away unmatched pairs, to optimize the 3D backbone $\boldsymbol{\Theta_P}$.

\textbf{Problem Formulation.}\quad Extended on the paradigm, our goal is to build a general 3D self-supervised learning framework, that affords a wide range of downstream 3D perception tasks like semantic segmentation, object detection, and panoptic segmentation. We seek to perform scene-level consistency of 2D-3D elements relying on coherent semantic cues provided by coupling the recent popular VFM with our multi-modality semantic prototypes. Thus, our self-supervised objective is more explicit  by replacing the frame-level consistency, which is prone to be to be influenced by spatio-temporal movement or scene changes.

\textbf{Approach Overview.}\quad Our framework is outlined in Fig. \ref{fig:framework}. CSC framework consists of two key components: (1) a VFM-assisted semantic prototype generation covering semantic categories for large-scale cross-views/scenes (Sec. \ref{sec:2d_3d_prototypes}) and (2) a coherent semantic consistency between superpoints and prototypes for 3D representation learning (Sec. \ref{sec:prototype-based-Constraint}). The main differences from the current paradigm are that i) we leverage the VFM to obtain semantic-aware superpixels with cross-view/scene associations and ii) we propose multi-modality semantic prototypes to mine coherent semantic consistency for general 3D representation learning. 

Formally, according to the VFM-assisted superpixels $\boldsymbol{\mathcal{S_{\text{2D}}}}$, we obtain the  semantic prototype features $\boldsymbol{\mathcal{F}_{\text{2D}}}$ \& $\boldsymbol{\mathcal{F}_{\text{3D}}}$  for 2D and 3D data. Then, coupling these features with semantic cues $\boldsymbol{\mathcal{C}_{\text{sem}}}$ from VFM, we maintain two separate modality prototypes $\boldsymbol{\mathcal{P}_{\text{2D}}}$ \& $\boldsymbol{\mathcal{P}_{\text{3D}}}$ , which represent coherent semantic representations for cross-scene objects with the same semantic. Moreover, to fully utilize two prototypes from heterogeneous space, we design a multi-modality prototype blending mechanism, which consists of modality-specific prototype projection and multi-modality prototype fusion modules, resulting in a mixed prototype $\boldsymbol{\mathcal{P}_{\text{mix}}}$ containing rich multi-modality information. Upon on the $\boldsymbol{\mathcal{P}_{\text{mix}}}$ and $\boldsymbol{\mathcal{F}_{\text{3D}}}$, we derive a coherent semantic consistency loss $\boldsymbol{\mathcal{L}_{\text{pro}}}$ to pushes close superpoint embeddings $\boldsymbol{\mathcal{F}_{\text{3D}}}$ to the mixed prototypes $\boldsymbol{\mathcal{P}_{\text{mix}}}$. Due to our cross-scene semantic prototypes, we achieve the scene-level semantic consistency for 3D representation learning. According to the observations of the universal improvement on three downstream tasks (Tab. \ref{tab:ann_eff_sem_seg}), our scene-level semantic consistency constraint endows the generalization of 3D representations to various scene perception tasks.

\subsection{VFM-Assisted Semantic Prototype Generation}
\label{sec:2d_3d_prototypes}
Let us start by generating the semantic-aware superpixel $\boldsymbol{\mathcal{S_{\text{2D}}}}$ from 2D vision foundation model, and then evolve two separate yet semantically aligned prototypes from complementary modalities.

\textbf{Suerpixel \& Superpoint Embeddings.}\quad
Firstly, we utilize the pre-calibrated pose information to project each point $\mathbf{P}_{k}$ onto a camera image $\mathbf{I}_{l}$. Then, we leverage a VFM, DINOv2 \cite{DINOv2} by default, to group visually similar regions into $Q$ superpixels $\boldsymbol{\mathcal{S_{\text{2D}}}}=\left\{\boldsymbol{S}^q_{\text{2D}} \mid q=1,\ldots, Q \right\}$, where $\boldsymbol{S_q^{\text{2D}}}$ denotes the group of pixels belonging to the $q$-th superpixel. Combining the 2D-3D mapping and superpixels $\boldsymbol{\mathcal{S}}_{\text{2D}}$, we can obtain the associated superpoints $\boldsymbol{\mathcal{S}_{\text{3D}}}=\left\{\boldsymbol{S}_{\text{3D}}^q \mid q=1,\ldots, Q \right\}$. Subsequently, pairing with pixel/point-wise features generated from the 2D/3D embedding networks, we are able to obtain superpixel and  superpoint embeddings, $\boldsymbol{\mathcal{F}_{\text{2D}}}=\left\{\boldsymbol{F}_{\text{2D}}^q \mid q=1,\ldots, Q \right\}$ and   $\boldsymbol{\mathcal{F}_{\text{3D}}}=\left\{\boldsymbol{F}^q_{\text{3D}} \mid q=1,\ldots, Q \right\}$, by averaging pooling the pixel and point features, where $\boldsymbol{F}_{\text{2D}}^q$ / $\boldsymbol{F}_{\text{3D}}^q$ is the $q$-th superpixel/superpoint embeddings. 

\textbf{Multi-Modality Prototype Generation.} \quad For all pairs of superpixels and superpoints from diverse scenes, we uniformly assign them with the semantic signs $\boldsymbol{\mathcal{C}_{\text{sem}}}$ shared all scenes. The $\boldsymbol{\mathcal{C}_{\text{sem}}}$ is obtained from the category-sensitive VFM, where we get the refined VFM on arbitrary semantic segmentation benchmark. Subsequently, according to the $\boldsymbol{\mathcal{C}_{\text{sem}}}$, we can group superpixel\&superpoint embeddings $\boldsymbol{\mathcal{F}_{\text{2D}}}$ \& $\boldsymbol{\mathcal{F}_{\text{3D}}}$ with the same semantic sign, to obtain the 2D\&3D semantic prototype features, $\boldsymbol{\mathcal{P}_{\text{2D}}}=\left\{\boldsymbol{P}_{\text{2D}}^t\mid t=1,\ldots,T\right\}$ \& $\boldsymbol{\mathcal{P}_{\text{3D}}}=\left\{\boldsymbol{P}_{\text{3D}}^t \mid t=1,\ldots,T\right\}$, by performing an averaging operation. The process of the multi-modality prototype generation can be expressed as follows:
\vspace{-0.5em}
\begin{equation}
\begin{aligned}
& \boldsymbol{P}_{\text{2D}}^t=\frac{1}{\left| \boldsymbol{\mathcal{C}}^{t}_{\text{sem}} \right|} \sum_{\boldsymbol{S}_{\text{2D}}^q = c^{t}} \boldsymbol{F}_{\text{3D}}^t,  \\
& \boldsymbol{P}_{\text{3D}}^t=\frac{1}{\left| \boldsymbol{\mathcal{C}}^{t}_{\text{sem}} \right|} \sum_{\boldsymbol{S}_{\text{3D}}^q = c^{t}} \boldsymbol{F}_{\text{2D}}^t,
\end{aligned}
\vspace{-0.5em}
\end{equation}
where $|\boldsymbol{\mathcal{C}}^{t}_{\text{sem}}|$ is the count of the superpixels with the same semantic sign $t$. The total number of semantic signs is $T$. In our experiments, we use mask2former-based DINOV2 \cite{DINOv2} as the mask network to generate $\boldsymbol{\mathcal{C}_{\text{sem}}}$, where the network is fine-tuned on the ADE20K dataset \cite{ADE20K} including $T=150$ semantic classes. 

\textbf{Dicussion.}\quad
 From the pioneering work SLidR \cite{sautier2022image} to the amazing study Seal \cite{liu2023segment}, the superpixels generation has transitioned from the non-learning segmentation algorithm (\emph{i.e.}, SLIC \cite{SLIC} in \cite{sautier2022image}) to the category-insensitive VFM (\emph{i.e.}, SAM \cite{SAM} in \cite{liu2023segment}). The driving force behind improving segmentation algorithms stems from the annoying  self-conflict challenge within contrastive-based self-supervised frameworks. To circumvent this impediment, Seal \cite{liu2023segment} first introduces category-insensitive VFMs (such as SAM \cite{SAM} and SEEM \cite{SEEM}) to improve the quality of superpixels, significantly reducing the self-conflict issue between over segmentation and semantic consistency in each image. Attracted by the dramatic performance improvement from incorporating VFMs, we also adopt the powerful VFM and further devise our scene-level consistency pre-training framework. Compared to Seal, we exploit believable and consistent semantic cues provided by category-sensitive VFMs to alleviate self-conflict across all scenes.

\subsection{Coherent Semantic Consistency}
\label{sec:prototype-based-Constraint}
Based on the VFM-assisted semantic prototype, we propose a coherent semantic consistency to alleviate the challenge of cross-scenes self-conflict and conduct scene-level semantic regularization for 3D representation learning.
To achieve an ideal 3D backbone, it is imperative to explore the information-rich multi-modality prototypes. However, although the two prototypes of different modalities have been semantically aligned, they do not lie in a uniform feature space. To reduce this gap, we design a multi-modality prototype blending module consisting of modality-specific prototype projection and multi-modality prototype fusion sub-modules. By parallel performing feature projection on each modality prototype followed by the fusion of multi-modality prototypes, the blending module will generate information-rich hybrid prototypes $\boldsymbol{\mathcal{P}_{\text{mix}}}$. Considering both $\boldsymbol{\mathcal{P}_{\text{mix}}}$ and $\boldsymbol{\mathcal{P}_{\text{3D}}}$, we could achieve the scene-level semantic contrastive loss $\boldsymbol{\mathcal{L}_{\text{pro}}}$ to endow the pre-trained 3D backbone with the ability to stable semantic discrimination on complex and dynamic large-scale autonomous driving scenes.In the next, we would illustrate each component in detail. 

\textbf{Multi-Modality Prototype Blending}\quad
The MMPB module sequentially achieves feature alignment and fusion of heterogeneous modality prototypes via the modality-specific prototype projection and multi-modality prototype fusion modules. The two modules are defined as follows:
\begin{enumerate}
\setlength{\itemindent}{0.0cm}
    \item 
    { \textbf{Modality-Specific Prototype Projection.}\quad Given the $\boldsymbol{\mathcal{P}_{\text{2D}}}$ and $\boldsymbol{\mathcal{P}_{\text{3D}}}$, several linear layers are employed in parallel to implicitly project the feature space of different modality to uniform one, resulting the updated 2D prototypes $\boldsymbol{\mathcal{\Bar{P}}_{\text{2D}}}=\left\{\boldsymbol{\bar{P}}_{\text{2D}}^t \mid t=1,\ldots,T\right\}$ and  3D prototypes $\boldsymbol{\mathcal{\Bar{P}}_{\text{3D}}}=\left\{\boldsymbol{\bar{P}}_{\text{3D}}^t \mid t=1,\ldots,T\right\}$. This computation can be expressed as:
    \vspace{-0.5em}
    \begin{equation}
    \begin{aligned}
    \{\boldsymbol{P}_{\text{2D}}^{1},\ldots,\boldsymbol{P}_{\text{2D}}^{T} \} \xrightarrow[]{\text{Linear Layers}} \{\boldsymbol{\Bar{P}}_{\text{2D}}^{1},\ldots,\boldsymbol{\Bar{P}}_{\text{2D}}^{T} \}, \\
    \{\boldsymbol{P}_{\text{3D}}^{1},\ldots,\boldsymbol{P}_{\text{3D}}^{T} \} \xrightarrow[]{\text{Linear Layers}} \{\boldsymbol{\Bar{P}}_{\text{3D}}^{1},\ldots,\boldsymbol{\Bar{P}}_{\text{3D}}^{T} \}. \\
    \end{aligned}
    \vspace{-0.5em}
    \end{equation} 
}
    \item 
    {\textbf{Multi-Modality Prototype Fusion.}\quad Then, we fuse the prototypes from two modality prototypes $\boldsymbol{\mathcal{\Bar{P}}_{\text{2D}}}$ \& $\boldsymbol{\mathcal{\Bar{P}}_{\text{3D}}}$ that are both semantic category and feature space aligned, resulting in  mixed prototypes $\boldsymbol{\mathcal{\Bar{P}}_{\text{mix}}}=\left\{\boldsymbol{\bar{P}}_{\text{mix}}^t \mid t=1,\ldots,T\right\}$. The computation is:
    \vspace{-0.5em}
    \begin{equation}
    \small
    \begin{aligned}
    \{\boldsymbol{\Bar{P}}_{\text{2D}}^{1}, \ldots,\boldsymbol{\Bar{P}}_{\text{2D}}^{T}, \boldsymbol{\Bar{P}}_{\text{3D}}^{1},\ldots,\boldsymbol{\Bar{P}}_{\text{3D}}^{T} \} \xrightarrow[]{\text{Linear Layers}} \left\{\boldsymbol{P}_{\text{mix}}^{1},\ldots,\boldsymbol{P}_{\text{mix}}^{T} \right\}.
    \end{aligned}
    \vspace{-0.5em}
    \end{equation}
}
\end{enumerate}

Following the MMPB, we obtain the blended prototypes $\boldsymbol{\mathcal{{P}}_{\text{mix}}}$ consist of the complementary multi-modality information. Thus, if using $\boldsymbol{\mathcal{{P}}_{\text{mix}}}$, the resulting 3D backbone will equip the comprehensive discrimination from both 2D image modality and 3D lidar modality. 

\textbf{Prototype-based Loss.}\quad
We propose a scene-level semantic contrastive loss $\boldsymbol{\mathcal{L}_{\text{proto}}}$ between $\boldsymbol{\mathcal{P}_{\text{3D}}}$ and $\boldsymbol{\mathcal{{P}}_{\text{mix}}}$, to endow the pre-trained 3D backbone with the ability to coherence semantic discrimination on complex and dynamic large-scale autonomous driving scenes. Formally, the prototype-based contrastive loss $\boldsymbol{\mathcal{L}_{\text{pro}}}$ is defined as follows:
\vspace{-0.5em}
\begin{equation}
\begin{aligned}
\boldsymbol{\mathcal{L}_{\text{pro}}}=-\log \frac{\exp \left(\left\langle \boldsymbol{F}_{\text{3D}}, \boldsymbol{P}_{\text{mix}}^{+}\right\rangle / \tau_{pro}\right)}{\sum_{i=0}^{|\boldsymbol{\mathcal{C_\text{sem}}}|} \exp \left(\left\langle \boldsymbol{F}_{\text{3D}}, \boldsymbol{P}_{\text{mix}}^{i}\right\rangle / \tau_{pro}\right)},
\end{aligned}
\vspace{-0.5em}
\end{equation}
where $\left\langle \cdot, \cdot \right\rangle$ denotes the scalar production. The sign $\boldsymbol{P}_{\text{mix}}^{+}$ is the positive prototype of superpoint embedding $\boldsymbol{F}_{\text{3D}}$. The symbol $\tau_{pro}$ is a temperature hyper-parameter.

\textbf{Discussion.}\quad
Here, we argue that using coherent semantic cues from VFM is much better than the commonly adopted traditional cluster algorithm in the benefits for multi-modality prototypes fusion. Mostly self/unsupervised methods \cite{cclnet2023,growsp} leverage an unsupervised clustering algorithm, such as K-Means and DBSCAN, to produce the their prototypes. However, these methods all face a common challenge that requires manually adjusting clustering parameters based on the distribution of a specific dataset. In addition to the problem of hand-crafted parameters, there existing an other tricky challenge in the multi-modality self-supervised pre-training task, which is the alignment problem across different modalities. Fortunately, the introduction of class-sensitive VFM is able to bypass the above trouble challenges without any additional effort. 

\subsection{Loss Functions} 
Overall, our pre-training framework consists of two losses. \textbf{(1)} We utilize the common superpixel-superpoint contrastive loss $\boldsymbol{\mathcal{L}_{\text{sp}}}$ \cite{sautier2022image} to optimize our backbones:
\vspace{-0.5em}
\begin{equation}
\begin{aligned}
\boldsymbol{\mathcal{L}_{\text{sp}}}=-\sum_{i=0}^{Q} \log \frac{\exp \left(\left\langle\boldsymbol{F}_{\text{3D}}^{i}, \boldsymbol{F}_{\text{2D}}^{i}\right\rangle / \tau_{sp}\right)}{\sum_{j=0}^{Q} \exp \left(\left\langle\boldsymbol{F}_{\text{3D}}^i, \boldsymbol{F}_{\text{2D}}^{j}\right\rangle / \tau_{sp}\right)},
\label{eq:sp}
\end{aligned}
\vspace{-0.5em}
\end{equation}
where $i$-th superpoint feature $\boldsymbol{F}_{\text{3D}}^{i}$ and $i$-th superpixel feature $\boldsymbol{F}_{\text{2D}}^{i}$ are matched according to the calibration between point cloud frames and the related surround images. \textbf{(2)} We leverage the proposed prototype-based loss $\boldsymbol{\mathcal{L}_{\text{pro}}}$ to provide the global semantic consistency for 3D representation learning. Our total loss is given by:
\vspace{-0.5em}
\begin{equation}
\begin{aligned}
\boldsymbol{\mathcal{L}_{\text {total}}}=\boldsymbol{\mathcal{L}_{\text{sp}}}+ \boldsymbol{\mathbf{1}}_{\{n > \lambda\}}\boldsymbol{\mathcal{L}_{\text{pro}}},
\end{aligned}
\vspace{-0.5em}
\end{equation}
where the indicator of $\boldsymbol{\mathbf{1}}_{\{n > \lambda\}}$ takes the value $1$ if $n > \lambda$ and 0 otherwise, where $n$ is the current training epoch and $\lambda$ is the hyper-parameter that controls the starting epoch of using $\boldsymbol{\mathcal{L}_{\text{pro}}}$. By default, $\lambda=5$, $\tau_{sp}=0.07$, and $\tau_{pro}=1.0$.
% \vspace{-1.4em}

\section{Experiments}
\label{sec:experiments}
In this section, we present the experimental results of three different 3D perception tasks, each implemented by the popular 3D backbone of its domain. Specifically, semantic segmentation implemented by MinkUNet  \cite{choy20194d} in Sec. \ref{sec:exp_sem_seg}, object detection implemented by VoxelNet \cite{zhou2018voxelnet} in Sec. \ref{sec:exp_obj_det}, and panoptic segmentation implemented by Cylinder3D \cite{cylinder3d} in Sec. \ref{sec:exp_pan_seg}.  Conveniently, we draw Tab. \ref{tab:three_ann_eff_tasks} to show the comprehensive  comparison of CSC with existing methods on three perception tasks with limited  labeling. In addition, we study the role of each component in Sec. \ref{sec:ablation_study}. Due to the limited space, visualization  and other experiments would be shown in the supplementary materials. 

\begin{table}[t]
\centering
\scalebox{0.71}{
\begin{tabular}{l | c | c }
        \toprule
        \multirow{2}{*}{\bf Method \& Year}
        & \multicolumn{2}{c}{\bf Semantic Segmentation}
        \\
        \cline{2-3}
        {}
        & \multicolumn{1}{c}{{\textcolor{gray}{\textbf 1\%{\ }(mIoU{\ })}}} \vline
        & \multicolumn{1}{c}{{\textcolor{gray}{\textbf 5\%{\ }(mIoU{\ })}}}
        \\ 
        \midrule
        \multicolumn{3}{l}{\textit{MinkUNet}}
        \\
        Random Init.
        & 30.3
        & 47.7
        \\
        SLidR,{\ }22 \cite{sautier2022image}
        & 38.2
        & 52.2
        \\
        ST-SLidR,{\ }23 \cite{mahmoud2023self}
        & 40.7
        & 54.6
        \\   
        TriCC,{\ }23 \cite{pang2023unsupervised}
        & 41.2
        & 54.1
        \\ 
        Seal,{\ }23 \cite{liu2023segment}
        & 45.8
        & 55.6
        \\ 
        \rowcolor{gray!20}
        Ours
        & \bf 47.0 \small{\textcolor{better}{(+1.2{\ }mIoU)}}
        & \bf 57.0 \small{\textcolor{better}{(+1.4{\ }mIoU)}}
        \\
        \toprule
        \multirow{2}{*}{\bf Method}
        & \multicolumn{2}{c}{\bf Object Detection}
        \\
        \cline{2-3}
        {}
        & \multicolumn{1}{c}{{\textcolor{gray}{\textbf 5\%{\ }(mAP{\ }/{\ }NDS{\ })}}} \vline
        & \multicolumn{1}{c}{{\textcolor{gray}{\textbf 20\%{\ }(mAP{\ }/{\ }NDS{\ })}}}
        \\
        \midrule
        \multicolumn{3}{l}{\textit{VoxelNet + CenterPoint}}
        \\
        Random Init.
        & 38.0\ /\ 44.3
        & 50.2\ /\ 59.7
        \\
        SLidR,{\ }22 \cite{sautier2022image}
        & 43.3\ /\ 52.4
        & 50.4\ /\ 59.9
        \\
        TriCC,{\ }23  \cite{pang2023unsupervised}
        & 44.6\ /\ \bf{54.4}
        & 50.9\ /\ \bf{61.3}
        \\
        \rowcolor{gray!20}
        Ours
        & \bf 45.3{\ }/{\ }54.2 \small{\textcolor{better}{(+0.9{\ }mAP)}}
        & \bf 51.9{\ }/{\ }61.3 \small{\textcolor{better}{(+1.0{\ }mAP)}}
        \\
        \toprule
        \multirow{2}{*}{\bf Method}
        & \multicolumn{2}{c}{\bf Panoptic Segmentation}
        \\
        \cline{2-3}
        {}
        & \multicolumn{1}{c}{{\textcolor{gray}{\textbf 1\%{\ }(PQ{\ }/{\ }SQ{\ }/{\ }RQ)}}} \vline
        & \multicolumn{1}{c}{{\textcolor{gray}{\textbf 5\%{\ }(PQ{\ }/{\ }SQ{\ }/{\ }RQ)}}}
        \\
        \midrule
        \multicolumn{3}{l}{\textit{Cylinder3D + Panoptic-PolarNet}}
        \\
        Random Init.
        & 15.3{\ }/{\ }62.6{\ }/{\ }20.4
        & 20.9{\ }/{\ }73.4/26.5
        \\
        SLidR,{\ }22 \cite{sautier2022image}
        & 16.3{\ }/{\ }65.7{\ }/{\ }21.4
        & 21.6{\ }/{\ }73.5{\ }/{\ }27.1
        \\
        \rowcolor{gray!20}
        Ours
        & \bf 19.3{\ }/{\ }74.5{\ }/{\ }24.6 \small{\textcolor{better}{(+3.0{\ }PQ)}}
        & \bf 23.1{\ }/{\ }76.9{\ }/{\ }28.5 \small{\textcolor{better}{(+1.5{\ }PQ)}}
        \\
        \bottomrule
\end{tabular}
}
\vspace{-0.5em}
\caption{On nuScenes, CSC is compared with current state-of-the-art methods in three downstream tasks with limited annotation. Obvious improvement in term of semantic segmentation, object detection, and panoptic segmentation could be found.} 
\label{tab:three_ann_eff_tasks}
\vspace{-1.5em}
\end{table}

\textbf{Datasets.} We pre-train all three models on nuScenes dataset, which is a large-scale autonomous driving dataset including 1,400,000 camera images as well as 90,00 Lidar sweeps across 1000 scenes. On nuScenes dataset, each point cloud keyframe is equipped with six calibrated surround images. During pre-training phase, we use the unlabeled RGB images and point clouds from 600 scenes to update backbones in our CSC, same as SLidR. About fine-tuning on three 3D perception tasks, we all conduct experiments on nuScenes, to evaluate the quality of pre-trained 3D backbone with various percentage annotations. The nuScenes dataset is also used in our fine-tuning for three perception task, to evaluate the annotation-efficient of the various pre-training methods under different percentages of labeling.

\begin{table*}[t]
\centering
\scalebox{0.9}{
\setlength{\tabcolsep}{3.2 pt}
\begin{tabular}{l | c | c c c c c c | c}
        \toprule
        \multirow{2}{*}{\bf Method}
        &  \multirow{2}{*}{\bf Venue}
        &  \multicolumn{6}{c}{\bf nuScenes} \vline
        &  \multicolumn{1}{c}{\bf KITTI}
        \\ 
        \cline{3-9}  
        {}
        & {}
        & {\textcolor{gray}{\bf LP}}
        & {\textcolor{gray}{\bf 1\%}}
        & {\textcolor{gray}{\bf 5\%}}
        & {\textcolor{gray}{\bf 10\%}}
        & {\textcolor{gray}{\bf 25\%}}
        & {\textcolor{gray}{\bf 100\%}}
        & {\textcolor{gray}{\bf 1\%}}
        \\ 
        \midrule
        Random Init.
        & N/A
        & 8.1
        & 30.3
        & 47.7
        & 56.6
        & 64.8
        & 74.2
        & 39.5
        \\
        Point Con. \cite{xie2020pointcontrast}
        & ECCV 2020
        & 21.9
        & 32.5
        & - 
        & 57.1
        & -
        & 74.3
        & 41.1
        \\
        Depth Con. \cite{zhang2021self}
        & ICCV 2021
        & 22.1
        & 31.7  
        & -
        & 57.3
        & -
        & 74.1
        & 41.5
        \\
        PPKT \cite{liu2021learning}
        & arXiV 2021
        & 35.9
        & 37.8
        & 51.7
        & 59.2
        & 66.8
        & 73.8
        & 44.0
        \\
        SLidR \cite{sautier2022image}
        & CVPR 2022
        & 38.0
        & 38.2
        & 52.2
        & 58.8
        & 66.2
        & 74.6
        & 44.6
        \\
        ST-SLidR \cite{mahmoud2023self}
        & CVPR 2023
        & 40.4
        & 40.7
        & 54.6
        & 60.7
        & 67.7
        & 75.1
        & 44.7
        \\ 
        TriCC \cite{pang2023unsupervised}
        & CVPR 2023
        & 38.0
        & 41.2
        & 54.1
        & 60.4
        & 67.6
        & 75.6
        & 45.9
        \\ 
        Seal \cite{liu2023segment}
        & NeurIPS 2023
        & 44.9
        & 45.8
        & 55.6
        & 62.9
        & 68.4
        & 75.6
        & 46.6
        \\ 
        \rowcolor{gray!20}
        Ours
        & -
        & \bf 46.0 \small{\textcolor{better}{(+1.1)}}
        & \bf 47.0 \small{\textcolor{better}{(+1.2)}}
        & \bf 57.0 \small{\textcolor{better}{(+1.4)}}
        & \bf 63.3 \small{\textcolor{better}{(+0.4)}}
        & \bf 68.6 \small{\textcolor{better}{(+0.2)}}
        & \bf 75.7 \small{\textcolor{better}{(+0.1)}}
        & \bf 47.2 \small{\textcolor{better}{(+0.6)}}
        \\
        \bottomrule
\end{tabular}
}
\vspace{-0.5em}
\caption{Results (mIoU) of different pre-training methods on  semantic segmentation fine-tuning. On nuScenes, we use 100\% annotated scans for linear probing and 1\%, 5\%, 10\%, 25\%, 100\% annotation for fine-tuning. In addition, we use 1\% labels for fine-tuning on SemanticKITTI.} 
\label{tab:ann_eff_sem_seg}
\vspace{-1.5em}
\end{table*}

\textbf{Pre-training Details.}  Due to variances in network architectures, various 3D networks require different configurations in pre-training. For MinkUNet, we use the SGD optimizer with the 2.0 initial learning rate and a cosine annealing learning rate scheduler with a total of 50 epochs. The pre-training configuration of VoxelNet is similar to that of MinkUNet, the difference is that the initial learning rate is 0.01. As for Cylinder3D, we use the Adam optimizer of 0.001 initial learning rate and also employ the cosine annealing learning rate scheduler with a total of 15 epochs. All pre-trained 3D backbones are done with 2 RTX A6000 with a batch size 16.

\subsection{Annotation-Efficient Semantic Segmentation}
\label{sec:exp_sem_seg}
In this section, we measure the information of semantics learned by the 3D representations using various self-supervised frameworks. Overall, we compared CSC with the state-of-the-art methods on two benchmark datasets. In details, we evaluate the fine-tuned  semantic segmentation performance of pre-trained 3D backbone on nuScenes \cite{caesar2020nuscenes} and SemanticKITTI \cite{behley2019semantickitti} datasets.

Following SLidR \cite{sautier2022image}, we fine-tune the pre-trained 3D backbone using various percentage point cloud subsets with 1\%, 5\%, 10\%, 25\%, and 100\% of annotations for nuScenes and 1\% for SemanticKITTI. Meanwhile, we conduct a linear evaluation using 100\% annotations, which trains only a linear head and freezes other layers of the 3D backbone, to investigate the generalizability of representations learned via self-supervised learning without task-specific fine-tuning. We report the metric of mean Iou (mIoU) to evaluate various methods.

In Tab. \ref{tab:ann_eff_sem_seg}, we show the comparison of the previous methods with CSC. It is evident that the 3D backbone with pre-trained parameters derived from arbitrary 3D self-supervised pre-training framework substantially outperforms the random initialized one. Compared with the current state-of-the-art method Seal \cite{liu2023segment} on nuScenes, our CSC provides significant mIoU improvements of +1.1\% for linear probing, +1.2\% and +1.4\% for 1\% and 5\% few-shot fine-tuning settings, respectively. In addition, CSC also achieves better generalization of +0.6\% boosting on the out-of-distribution annotation-efficient semantic segmentation in the SemanticKITTI. Compared to Seal, who only utilizes a class-insensitive VFM for each individual image to alleviate the self-conflict problem, our CSC presents a better 3D network with strong discriminative power. This suggests the importance of embracing the coherent semantic cues from the class-sensitive VFM and the scene-level semantic consistency in the pre-training phase. Due to page limitations, we show the average per-class performance of 1\% annotation for detailed analysis in supplementary materials.

\begin{table}
\centering      
\scalebox{0.85}{
\begin{tabular}{l|c c|c c|c c}        
        \toprule        
        % \multicolumn{1}{l|}{Pretrain} 
        \multirow{3}{*}{\bf Method} 
        &  \multicolumn{6}{c}{\bf nuScenes}
        \\
        \cline{2-7}
        & \multicolumn{2}{c}{\textcolor{gray}{\bf 5\%}}  \vline 
        & \multicolumn{2}{c}{\textcolor{gray}{\bf 10\%}} \vline
        & \multicolumn{2}{c}{\textcolor{gray}{\bf 20\%}}
        \\    
        {}
        & mAP 
        & NDS 
        & mAP 
        & NDS 
        & mAP 
        & NDS 
        \\
        \midrule 
        \multicolumn{7}{l}{\textit{VoxelNet + CenterPoint}} 
        \\
        Random Init.
        & 38.0
        & 44.3
        & 46.9
        & 55.5
        & 50.2
        & 59.7
        \\
        Point Con. \cite{xie2020pointcontrast}
        &39.8
        &45.1 
        &47.7
        &56.0 
        &-
        &-
        \\        
        GCC-3D  \cite{liang2021exploring}
        &41.1
        &46.8 
        &48.4
        &56.7 
        &-
        &-
        \\        
        SLidR  \cite{sautier2022image}
        &43.3
        &52.4 
        &47.5
        &56.8 
        &50.4
        &59.9
        \\  
        TriCC \cite{pang2023unsupervised}
        &44.6
        &\bf 54.4 
        &48.9
        &58.1 
        &50.9
        & 60.3
        \\  
        \rowcolor{gray!20}
        Ours
        & \textbf{45.3}
        & 54.2
        & \textbf{49.3}
        & \textbf{58.3}
        & \textbf{51.9}
        & \textbf{61.3}
        \\ 
        \bottomrule    
        \multicolumn{7}{l}{\textit{VoxelNet + SECOND}}
        \\
        Random Init.
        &35.8
        &45.9
        &39.0
        &51.2
        &43.1
        &55.7
        \\
        SLidR  \cite{sautier2022image}
        &36.6
        &48.1 
        &39.8
        &52.1 
        &44.2
        &56.3
        \\ 
        TriCC  \cite{pang2023unsupervised}
        &37.8
        &\bf 50.0 
        &41.4
        &53.5 
        &45.5
        &57.7
        \\  
        \rowcolor{gray!20}
        Ours
        &\textbf{38.2}
        &49.4
        &\textbf{42.5}
        &\textbf{54.8}
        &\textbf{45.6}
        &\textbf{58.1}
        \\ 
        \bottomrule 
\end{tabular} 
}
\vspace{-0.5em}
\caption{Results (mAP and NDS) when fine-tuning the pre-trained backbones to object detection using two models (CenterPoint and SECOND) with 5\%, 10\%, and 20\% labels on nuScenes.}
\label{tab:ann_eff_obj_det}
\vspace{-1.5em}
\end{table}

\subsection{Annotation-Efficient Object Detection}
\label{sec:exp_obj_det}
In the vision system of autonomous driving, 3D object detection is a common and challenging task. Thus, we further evaluate the quality of our pre-trained lidar representation on this object-level task on the nuScenes. Following the previous works, we fine-tune the pre-trained 3D backbone with 5\%, 10\%, and 20\% of the labeled data, respectively. Moreover, we embed the pre-trained Cylinder3D into two detection models, CenterPoint and SECOND. We refer to the evaluation protocol of nuScenes \cite{caesar2020nuscenes} and report the mean average precision (mAP) and nuScenes detection score (NDS), where NDS is a weighted average of mAP that measures the quality of the detection in various terms.

In Tab.  \ref{tab:ann_eff_obj_det}, we compare the existing methods with our CSC. Compared to SLidR, the current SOTA method TriCC has gain the significantly improvement of 1.3\% and 1.2\% mAP in 5\% mAP annotations by using the temporal consistency loops on both detection models. Taking the excellent work TriCC as the comparison, our CSC achieves the improvement of 0.7\% mAP and 0.4\% mAP without explicit temporal consistency. Surprisingly, the growth in performance by our CSC is steady, even using more annotation.
% \vspace{-0.5em}

\subsection{Annotation-Efficient Panoptic Segmentation}
\label{sec:exp_pan_seg}
In this experiment, we compare the various pre-training methods for panoptic segmentation, which evaluates both semantic and instance recognition ability of the learned 3D backbone. To  our knowledge, CSC is the first pre-training framework that transferring the pre-trained 3D backbone, which absorbs the prior knowledge from 2D realm, to the more challenging and annotation-intensive panoptic segmentation. Considering fair comparisons, we refer to both the setting from the previous lidar-only pre-training method \cite{TARL} and the current state of development in the panoptic segmentation. Specifically, we employ the Panoptic-PolarNet \cite{zhou2021panoptic} with Cylinder3D \cite{cylinder3d}, just like the current supervised SOTA method \cite{Zhang_2023_ICCV}. About utilization rate of labels, we select the percentages of 1\%, 5\%, and 10\% to fine-tune the pre-trained network. Given the absence of existing methods conducted on panoptic segmentation, we treat SLidR as the primary comparison approach and using different 2D segmentation methods for superpixel generation. Overal, there are five experiments: random initialization, pre-training by SLidR and SLIC, pre-training by SLidR and Dinov2, pre-training by CSC and Dinov2, and pre-training by CSC and OneFormer. About the evaluation metrics, we report segmentation quality (SQ), recognition quality (RQ), and panoptic quality (PQ).

\begin{table}[t]
\centering
\scalebox{0.86}{
\begin{tabular}{l | c c c | c c c }
        \toprule
        \multirow{3}{*}{\bf Method} 
        & 
        \multicolumn{6}{c}{\bf nuScenes} 
        \\ 
        \cline{2-7} 
        & \multicolumn{3}{c}{\textcolor{gray}{\bf 1\%}}  \vline 
        & \multicolumn{3}{c}{\textcolor{gray}{\bf 5\%}}   
        \\
        {}
        & PQ  
        & SQ
        & RQ
        & PQ  
        & SQ
        & RQ
        \\ 
        \midrule
        Random Init.
        & 15.3
        & 62.6
        & 20.4
        & 20.9
        & 73.4
        & 26.5
        \\
        SLidR + SLIC
        & 16.3
        & 65.7
        & 21.4
        & 21.6
        & 73.5
        & 27.1
        \\
        SLidR + DINOV2
        & 17.6
        & 70.7
        & 22.7
        & 22.3
        & 75.1
        & 27.8
        \\ 
        \rowcolor{gray!20}
        Ours (default)
        & 19.3
        & 74.5
        & 24.6
        & 23.1
        & \bf 76.9
        & 28.5
        \\
        \rowcolor{gray!20}
        Ours + OneFormer
        & \bf 19.5
        & \bf 78.3
        & \bf 25.0
        & \bf 23.4
        & 75.7
        & \bf 28.8
        \\
        \bottomrule
\end{tabular}}
\vspace{-0.5em}
\caption{ Results (PQ, SQ, and RQ) when fine-tuning the pre-trained models to panoptic segmentation with 1\% and 5\% labels on nuScenes. Considering the absence of existing pre-training methods, besides random initialization and the default CSC settings, we additionally establish three sets of experiments: original SLidR with SLIC, SLidR with DINOV2, and CSC with OneFormer.} 
\label{tab:ann_eff_pan_seg}
\vspace{-2em}
\end{table}

Tab. \ref{tab:ann_eff_pan_seg} shows that multi-modality pre-training method is consistently better than the random initialization. With 1\% annotation, replacing the superpixel generation method from SLIC to Dinov2 and further adopting our CSC can improve the PQ metric by 1.3\% and 3.0\%, respectively, compared to the original SLidR. On top of CSC, replacing DINOv2 by other semantic segmentation networks, such as OneFormer, will result in a considerable performance gain of 3.2\%. Observing the changes in the SQ and RQ metrics of panorama segmentation across 1\% and 5\% ratios, we can find that the improvement brought from SLidR to our CSC is mainly in the SQ metrics (\emph{i.e.}, $65.7 \to 74.5$ and $73.5 \to 76.9$) while the improvement in the RQ metrics is relatively slight (\emph{i.e.}, $21.4 \to 24.7$ and $27.1 \to 28.5$). This phenomenon is consistent with previous results on semantic segmentation and object detection, \emph{i.e.}, our approach significantly improves 3D network's ability to recognize semantic categories while providing a limited increase in the ability to discriminate different instances with the same semantics.

\begin{table}[t]
\centering
\scalebox{0.78}{
\begin{tabular}{ c| c c c c | c c c c c}
\toprule
\multirow{2}{*}{\bf{\#}} 
& \multirow{2}{*}{\bf{SP}} 
& \multirow{2}{*}{\bf{VFM}} 
& \multirow{2}{*}{\bf{3D Pro.}} 
& \multirow{2}{*}{\bf{MMPB}} 
& \multicolumn{4}{c}{\bf{nuScene}}
\\
\cline{6-9}
& 
& 
& 
& 
& \textcolor{gray}{\bf{LP}} 
& \textcolor{gray}{\bf{1\%}} 
& \textcolor{gray}{\bf{5\%}} 
& \textcolor{gray}{\bf{10\%}} 
\\
\midrule
(1) 
& \checkmark 
& 
& 
& 
& 38.0
& 38.2
& 52.2
& 58.8
\\
(2) 
& \checkmark 
& \checkmark 
& 
& 
& 44.0
& 41.1
& 52.1
& 60.9
\\
(3) 
& \checkmark 
& \checkmark 
& \checkmark 
& 
& 44.0 
& 40.3 
& 53.3 
& 60.5 
\\
(4) 
& \checkmark 
& \checkmark
& \checkmark 
& \checkmark 
& \bf 46.0 
& \bf 47.0
& \bf 57.0
& \bf 63.3 
\\
\bottomrule
\end{tabular}
}
\vspace{-0.5em}
\caption{Ablation study of each component pre-trained and fine-tuned on nuScenes. \textbf{SP}: Superpixel-Superpoint contrastive loss $\boldsymbol{\mathcal{L}_{\text{sp}}}$. \textbf{VFM}: Vision foundation models. \textbf{3D Pro.}: The mixed prototypes is replace by the raw 3D prototypes in the $\boldsymbol{\mathcal{L}_{\text{pro}}}$. \textbf{MMPB}: Multi-Modality Prototype blending.}
\label{tab:ablation}
\vspace{-2.em}
\end{table}
\vspace{-1 em}

% \vspace{-1em}

\subsection{Ablation Study}
\label{sec:ablation_study}
\vspace{-1.0em}
We perform ablation experiments to examine the contribution of VFM for superpixel generation and multi-modality prototype blending (MMPB). We also investigate the impact of discarding MMPB to directly employ raw 3D prototypes for 3D backbone learning. All the ablation studies are conducted by 1\%, 5\%, and 10\% semantic segmentation fine-tuning and 100\% linear evaluation on nuScenes dataset.

Compared to the original SLidR \#(1) at 1\% labeling, introducing our VFM \#(2) and MMPB \#(4)  in turn, resulting in the steep rise of mIoU, from  $38.2 \to 41.4 \to 47.0$. This upward trend persists across various annotation ratios, however, it tends to decelerate with the increase in the number of labels. Interestingly, when comparing \#(3) with \#(4), we can observe that directly employing the 3D semantic prototypes assisted by the VFM for 3D representation learning results in a performance deterioration of 0.8\% mIoU, yet incorporating our proposed MPPB reverses this degradation and gains the promotion of 6.7\%.

% \textbf{Strong Pre-Training Baseline.}
% Through comprehensive evaluations across three mainstream perception tasks and ablation study, these results indicate that our CSC can achieve the general 3D backbone via exploring the coherent semantic cues from VFM. More specifically, our pre-trained 3D backbone consistent surpasses the existing state-off-the-art approaches in three downstream perception tasks using task-specific 3D backbone. This experimental results demonstrate that our framework both derives the universal 3D representation with strong generalizability across various tasks, and applicable to a wide range of networks in large-scale outdoor scenes.
\vspace{-1.5em}
% \vspace{-0.5em}
\section{Conclusion}
\vspace{-0.5em}
In this paper, we study the multi-modality 3D pre-training task from the drawbacks of the existing methods, including the self-conflict of cross-scene semantic segments and the absence of building the global semantic units. Our CSC performs scene-level semantic consistency via the combination VFM-assisted semantic cues and multi-modality semantic prototypes. Firstly, obtain the coherent semantic superpixels based on the VFM and use the semantics to generate prototypes for two modalities. Then,  compute the unified prototypes by modality-specific prototype projection with multi-modality prototype blending. Thereby, we can achieve the cluster contrastive loss between 3D superpoint features and mixed prototypes for learning universal 3D representation. Extensive experiments show that our method delivers state-of-the-art results on semantic segmentation, object detection, and panoptic segmentation.

\vspace{-0.5em}

\noindent
\textbf{Acknowledgment.} This work is supported by the National Natural Science Foundation of China No.62302167, U23A20343, 62222602, 62206293, 62176224, 62106075, and 62006139, Shanghai Sailing Program (23YF1410500), Natural Science Foundation of Shanghai (23ZR1420400), Science and Technology Commission of Shanghai No.21511100700, Natural Science Foundation of Chongqing, China (CSTB2023NSCQ-JQX0007, CSTB2023NSCQ-MSX0137), CCF-Tencent Rhino-Bird Young Faculty Open Research Fund (RAGR20230121), CAAI-Huawei MindSpore Open Fund, and Development Project of Ministry of Industry and Information Technology (Grant Number: ZTZB.23-990-016).
% \section{}

% xzx
% Finally, we achieve the 3D% scene-level semantic consistency to pre-train 3D backbone.% In this paper, we study the multi-modality 3D pre-training task from the perspective of performing scene-level semantic consistency through VFM-assisted semantic cues and multi-modality semantic prototypes. Our CSC  obtain the coherent semantic superpixels from the class-sensitive VFM and  utilizes these signs to build semantic prototypes for two heterogeneous modalities. Then, we design the MMPB to blend these unaligned prototypes by modality-specific prototype projection and multi-modality prototype blending. Finally, we achieve the scene-level semantic consistency to pre-train 3D backbone. Extensive experiments show that our method delivers state-of-the-art results on three downstream tasks, semantic segmentation, object detection, and panoptic segmentation.% coherent semantic sings from the class-sensitive VFM 

\label{sec:conclusion}

{
    \small
    \bibliographystyle{ieeenat_fullname}
    \bibliography{main}
}

% \input{X_suppl}

\end{document}